\documentclass[10pt,twocolumn,letterpaper]{article}

\usepackage[pagenumbers]{cvpr} 
\usepackage[accsupp]{axessibility}  

\makeatletter
\DeclareRobustCommand\onedot{\futurelet\@let@token\@onedot}
\def\@onedot{\ifx\@let@token.\else.\null\fi\xspace}

\def\eg{\emph{e.g}\onedot} \def\Eg{\emph{E.g}\onedot}
\def\ie{\emph{i.e}\onedot}

\makeatother

\usepackage{graphicx}
\usepackage{amsmath}
\usepackage{amssymb}
\usepackage{booktabs}
\usepackage{multirow}
\usepackage{float}
\usepackage{subcaption}
\usepackage{tikz}

\usepackage[pagebackref,breaklinks,colorlinks,allcolors=cvprblue]{hyperref}
\usepackage[capitalize]{cleveref}
\crefname{section}{Sec.}{Secs.}
\Crefname{section}{Section}{Sections}
\Crefname{table}{Table}{Tables}
\crefname{table}{Tab.}{Tabs.}

\def\supmat{Sup. mat.}

\usepackage{pifont}
\newcommand{\cmark}{\ding{51}}%
\newcommand{\xmark}{\ding{55}}%
\newcommand{\parsection}[1]{\noindent\textbf{#1:}}

\usepackage{nicematrix}
\definecolor{Grey}{RGB}{100, 100, 100}
\definecolor{CornflowerBlue}{RGB}{100, 149, 237}
\definecolor{YellowGreen}{RGB}{154, 205, 50}
\definecolor{Apricot}{RGB}{251, 206, 177}

\newcommand{\dashedbox}[1]{%
    \tikz[baseline=(X.base)] 
        \node[draw=black, dashed, rectangle, inner sep=2pt, line width=0.4mm, rounded corners] (X) {#1};%
}

\definecolor{cvprblue}{rgb}{0.21,0.49,0.74}

\def\paperID{xx} 
\def\confName{CVPR}
\def\confYear{2025}

\title{Exploring Semi-Supervised Learning for Online Mapping}

\author{Adam Lilja$^{1,2}$ \,\,\,\,\,\, Erik Wallin$^{1,3}$ \,\,\,\,\,\, Junsheng Fu$^{2}$ \,\,\,\,\,\, Lars Hammarstrand$^{1}$
\\
\normalsize$^1$Chalmers University of Technology \hspace{0.8cm} $^2$Zenseact \hspace{0.8cm} $^3$Saab AB \hspace{0.8cm}\\
{\tt\small \{firstname.lastname\}@\{zenseact.com, chalmers.se, saabgroup.com\}}
}

\begin{document}
\twocolumn[{
\maketitle
\begin{center}
    \captionsetup{type=figure}
    \includegraphics[width=\linewidth, trim={0mm, 8mm, 0mm, 2mm}, clip]{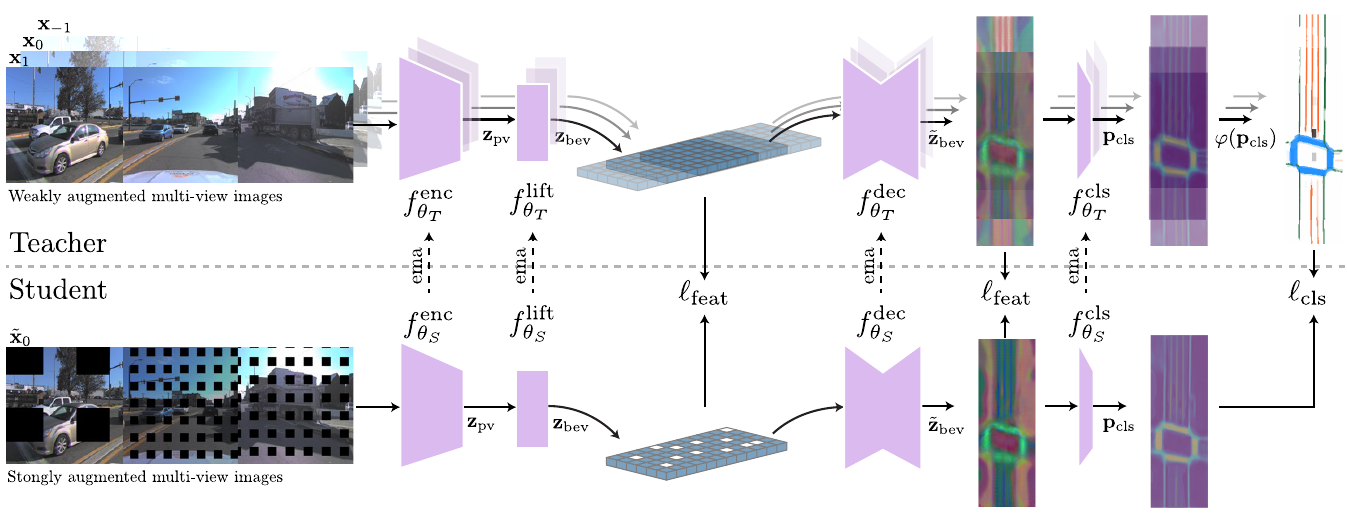}
    \vspace{-2mm}
    \captionof{figure}{
    Semi-supervised learning framework for online mapping with temporal pseudo-label fusion. 
    Our approach adapts the Teacher-Student paradigm to the online mapping domain while introducing a novel temporal fusion mechanism that aggregates pseudo-labels across multiple frames. 
    We enable label-efficient learning from largely unlabelled datasets and strong generalization to unseen cities.
    }
    \label{fig:main-arch}
\end{center}
}]

\begin{abstract}
The ability to generate online maps using only onboard sensory information is crucial for enabling autonomous driving beyond well-mapped areas. 
Training models for this task -- predicting lane markers, road edges, and pedestrian crossings -- traditionally require extensive labelled data, which is expensive and labour-intensive to obtain. 
While semi-supervised learning (SSL) has shown promise in other domains, its potential for online mapping remains largely underexplored. 
In this work, we bridge this gap by demonstrating the effectiveness of SSL methods for online mapping. 
Furthermore, we introduce a simple yet effective method leveraging the inherent properties of online mapping by fusing the teacher's pseudo-labels from multiple samples, enhancing the reliability of self-supervised training. 
If 10\% of the data has labels, our method to leverage unlabelled data achieves a 3.5x performance boost compared to only using the labelled data. 
This narrows the gap to a fully supervised model, using all labels, to just 3.5 mIoU.
We also show strong generalization to unseen cities. 
Specifically, in Argoverse 2, when adapting to Pittsburgh, incorporating purely unlabelled target-domain data reduces the performance gap from 5 to 0.5 mIoU.
These results highlight the potential of SSL as a powerful tool for solving the online mapping problem, significantly reducing reliance on labelled data.
\end{abstract}
    
\vspace{-7mm}
\section{Introduction}
\label{sec:introduction}
Safe autonomous navigation requires accurately perceiving both dynamic traffic participants and static road infrastructure.
Relying solely on pre-made offline High-Definition (HD) maps for the latter comes with difficulties, such as accurate localisation within the map, and the costs associated with building and keeping the maps up to date. 
The latter is significant for scaling Autonomous Driving (AD) beyond limited, well-mapped areas.
Online mapping is an alternative and complementary approach, where onboard sensory data is used to produce a local up-to-date map of surrounding lane markers, road edges, and pedestrian crossings in real-time \cite{li2022hdmapnet, liao2023maptr}.

Recent studies have shown that learning-based online mapping algorithms struggle to generalise to regions outside their training set, suggesting available data is insufficient for learning this complex task \cite{lilja2024localization, lindstrom2024nerfs, qin2023unifusion}.
While collecting large amounts of AD data is relatively straightforward, the bottleneck lies in the effort labelling the data. 
Creating the necessary labels, \eg 3D polylines for each lane divider, is generally time-consuming and costly. 
Therefore, we are interested in approaches utilizing a small amount of labelled data and a vast amount of unlabelled data to increase the generalisability in the online mapping task.

Other domains, such as image classification \cite{tarvainen2017mean} and semantic segmentation \cite{french2019semi}, have addressed this problem setting through Semi-Supervised Learning (SSL) and shown strong capabilities.
Recently, the interest in SSL for object detection has also gained traction and shown promising results \cite{liu2021unbiased, xu2021end}.
Some SSL techniques have been successfully adapted to the closely related domain of Bird's Eye View (BEV) semantic segmentation of the combination of dynamic classes such as vehicles and pedestrians, and static classes including vegetation and drivable areas.
However, the focus has been on \emph{monocular} BEV segmentation where the teacher-student setup \cite{tarvainen2017mean} has shown effectiveness \cite{zhu2024semibevseg,gao2022s2g2}.
New challenges arise when multiple cameras are used as input, as creating 3D-consistent augmentations in this context proves difficult.
For instance, the conjoint-rotation technique \cite{zhu2024semibevseg} is incompatible with our problem setup.
Concurrent work \cite{ishikawa2024pct} shows that randomly disabling camera feeds and applying BEV feature dropout as student augmentation techniques effectively enhance performance in a multi-camera setup.
Despite these advances, there remains a gap in understanding how SSL components specifically interact in the context of online mapping, where the focus is exclusively on static classes rather than the mixed dynamic-static scenario addressed in prior work.

In this work, we identify key SSL approaches that excel in online mapping, both for learning with limited labels and for adapting models to new cities.
As online mapping focuses solely on static classes, it allows us to exploit temporal consistency across frames.
Building on this insight, we propose a label-efficient SSL method, illustrated in \cref{fig:main-arch}, that adapts the Teacher-Student paradigm to online mapping while introducing a novel temporal fusion mechanism. 
This mechanism aggregates pseudo-labels across multiple frames, improving their reliability and reducing noise.
By leveraging large-scale unlabelled datasets, our approach significantly reduces dependence on labelled data while enabling strong generalization to unseen cities.

To summarize, we make four key contributions:
1) a comprehensive analysis of prevalent SSL techniques for online mapping, demonstrating which existing methods can significantly enhance performance.
2) combine SSL components into a robust method that maintains high performance levels across varying volumes of labelled data. 
3) propose that the teacher model integrates information across multiple samples to improve pseudo-label accuracy.
4) demonstrate strong performance in domain adaptation when trained in one group of cities and evaluated in another.
\section{Related Work}
\label{sec:related-work}
We explore semi-supervised learning and domain adaptation techniques specifically applied to online mapping. 
This section examines relevant prior work in these fields.

\subsection{Semi-Supervised Learning}
\label{sec:related-ssl}
In SSL, costly labelled data is combined with inexpensive unlabelled data to train a model.
There are two main approaches to this problem: single-stage training, where automatic ground truth labels are generated during training, \ie using predictions from a single model or an ensemble of models as pseudo-labels \cite{sohn2020fixmatch, zhang2021flexmatch,li2022pseco}, or a two-stage process starting with self-supervised pre-training followed by finetuning on the labelled data \cite{chen2020simple, he2022masked, caron2021emerging, ljungbergh2025gasp}. 
This paper concentrates on the single-stage joint training approach, exploring critical design decisions and their impact on the performance of semi-supervised online mapping.

SSL has a rich history in image classification, with the teacher-student framework \cite{tarvainen2017mean} being particularly influential. 
This framework involves a student network supervised by predictions from a teacher network. 
The teacher network shares the same architecture as the student but updates its weights as an exponential moving average (EMA) of the student's weights.
ReMixMatch \cite{berthelot2019remixmatch} further feeds the teacher weakly augmented inputs, while the student receives strongly augmented inputs. 
Data augmentation is crucial in SSL \cite{zhao2023augmentation, french2019semi, VERMA202290} and these may differ from those suitable for fully supervised settings \cite{zhao2023augmentation}.
Commonly used augmentation strategies include RandAugment \cite{cubuk2020randaugment}, CutOut \cite{devries2017improved}, CutMix \cite{yun2019cutmix}, and feature dropout \cite{yang2023revisiting}.
%

FixMatch \cite{sohn2020fixmatch} further popularized \emph{thresholding}, a technique where the student is supervised only with high-confidence pseudo-labels generated by the teacher.
Numerous follow-up studies continue to demonstrate the strengths of this method \cite{yang2023revisiting, wang2022freematch}.
Another line of work in SSL involves adding self-supervised auxiliary tasks to improve performance, such as predicting the parameters of stochastic image transformations applied to unlabelled images \cite{EnAET, Zhai_2019_ICCV} or adding similarity loss of the features between the teacher and student models \cite{chen2021exploring, wallin2022doublematch}.

Research in SSL has also been extended to BEV segmentation, considering a variety of classes such as car, truck, road, walkway, and vegetation \cite{gao2022s2g2, zhu2024semibevseg, gosala2023skyeye, gosala2024letsmap}. 
The teacher-student approach has, for instance, been adapted to monocular BEV segmentation, utilizing MSE losses for probability and feature similarity \cite{zhu2024semibevseg, gao2022s2g2}.
Unlike in Perspective View (PV) semantic segmentation, where a variety of augmentations are permissible, the augmentations applied in BEV segmentation must maintain 3D consistency. 
\Eg, methods such as randomly cropping and resizing the PV input image for BEV segmentation can break geometric consistency in the BEV plane. 
One augmentation with geometric consistency is the conjoint rotation introduced in \cite{zhu2024semibevseg} where both images and BEV pseudo-labels are rotated around the upward axis.
However, this technique is limited to monocular camera setups, as any augmentation affecting the 3D geometry must maintain 3D consistency across all cameras.

Our setup closely resembles the concurrent work of PCT \cite{ishikawa2024pct}, which proposes SSL on multi-view images using a teacher-student framework inspired by \cite{yang2023revisiting}. 
They introduce camera dropout augmentation, where one or more camera images are omitted for the student model, and leverage BEV feature dropout. 
Additionally, they use pre-trained models to generate PV semantic segmentation and depth pseudo-labels. 
In contrast, our work focuses on SSL methods that do not rely on additional pre-trained networks for pseudo-labeling. 
Another key distinction is that, unlike BEV segmentation, which includes dynamic object classes, online mapping is concerned solely with stationary elements.

\subsection{Domain Adaptation}
Unsupervised domain adaptation (UDA) is closely related to SSL, and throughout this paper, we will use the terms interchangeably. 
In UDA, the unlabelled data comes from a target domain that differs from the labelled source domain.
The goal of UDA is to achieve strong performance in unseen environments, which is particularly critical for online mapping, where models must perform well in cities they have never been trained on -- a problem often overlooked in prior work \cite{lilja2024localization}.
Similar to SSL, many UDA techniques rely on self-training with a teacher-student setup and have also demonstrated strong results across various UDA settings \cite{zou2019confidence}.
A dominant approach for perspective view semantic segmentation in the image plane for AD involves augmenting data by mixing crops of labelled and unlabelled samples \cite{tranheden2021dacs, chen2024transferring}.
However, such augmentations are poorly suited for online mapping, as they fail to maintain consistency across multiple camera views, a key requirement in our setting.
In this work, we focus on self-training for UDA in online mapping, emphasizing its importance in adapting to unseen cities -- a fundamental challenge for real-world deployment.

\subsection{Online Mapping}
Recently, significant progress has been made in the field of online mapping, leading to the development of two primary approaches: segmentation-based methods \cite{xie2022m, dong2023superfusion, li2022hdmapnet, Peng_2023_WACV} and vector-based methods \cite{liu2023vectormapnet, liao2023maptr, liao2023maptrv2, chen2024maptracker} where the key difference lies in the output representation. 
Segmentation-based methods rasterize a region from a Bird's Eye View and classify each cell, typically using a CNN~\cite{peng2023bevsegformer,dong2023superfusion,li2022hdmapnet}. 
The vector-based methods predict vectors with connected points, representing a polyline for each object, typically using a DETR-like \cite{carion2020endtoend} transformer decoder \cite{liao2023maptr, liao2023maptrv2, zhang2023online}.

Both approaches encode input images using a neural network and transform the features into BEV using a core technique known as \emph{lifting}. 
For instance, LSS \cite{philion2020lift} learns a categorical depth distribution for each pixel and use it to weigh how much the corresponding PV feature should influence the corresponding BEV-cell. 

Previous works have highlighted a performance drop as the domain gap between training and evaluation data increases. 
Particularly in the context of geographical distribution shifts \cite{lilja2024localization, qin2023unifusion}, and driving scenario shifts \cite{lindstrom2024nerfs}.
Our focus is on the effects of SSL approaches, and for this work, we choose a segmentation-based method, as its formulation aligns more naturally with existing SSL and UDA techniques.

\section{SSL for Online Mapping}
\label{sec:method}
Our work focuses on leveraging SSL to maximize the effectiveness of unlabelled data in online mapping.
In this context, we assume that we have a small set of labelled data ${\mathbf{X}_L=\{\mathbf{x_i}, \mathbf{y_i}\}_{i=1}^{n_l}}$, for which ground truth map $\mathbf{y}_i$ is available, and a large set of unlabelled images $\mathbf{X}_U = \{\mathbf{x_i}\}_{i=1}^{n_u}$ without ground truth annotations.
The latter is assumed to be collected in areas separate from the labelled set, but using the same camera setup. Further, the relative pose of the vehicle in the sequences is assumed to be known.

Using the labelled data, the student is trained in an ordinary supervised manner by minimizing a supervised loss ${\ell_\text{sup}(\mathbf{x_i}, \mathbf{y_i})}$, \eg, cross-entropy, DICE \cite{sudre2017generalised}, or Focal Loss \cite{lin2017focal}.
Effectively using the unlabelled data is more challenging as we do not have access to a ground truth map for this area, and several design choices must be made. 
To understand how to apply SSL methods in the context of online mapping, we explore the impact of these choices using a well-established segmentation-based online mapping architecture.
Figure \ref{fig:main-arch} shows an overview of the SSL training scheme, and in this section, we provide the details of the online mapping architecture, the prevalent SSL techniques we employed, and our proposed multi-step teacher fusion approaches.

\subsection{Online mapping architecture}
\label{sec:om-arch}
As we are mainly interested in the effects of SSL approaches, we choose to use a typical map segmentation architecture \cite{li2022hdmapnet, zhang2022beverse, Peng_2023_WACV} such as \cref{fig:main-arch} illustrates. 
The multi-view images, $\mathbf{x}$, are encoded by a neural network ${\mathbf{z}_\text{pv}=f^\text{enc}_\theta(\mathbf{x})}$ and lifted to BEV features as ${\mathbf{z}_{\text{bev}}=f^{\text{lift}}_\theta(\mathbf{\mathbf{z}_\text{pv}})}$. 
Then, another neural network decodes the BEV features ${\tilde{\mathbf{z}}_\text{bev}=f^{\text{dec}}_\theta(\mathbf{z}_{\text{bev}})}$.
Final class probabilities are predicted through a classification head ${\mathbf{p}_\text{cls}=f^\text{cls}_\theta(\tilde{\mathbf{z}}_{\text{bev}})}$, where ${\mathbf{p}_\text{cls}}$ comprise the independent probability of each map grid cell containing a lane delimiter, road edge, or pedestrian crossing.
As such, the online map is treated as a multi-label semantic segmentation problem. 
The complete architecture can be described as ${\mathbf{p}_\text{cls} = F_{\theta}(\mathbf{x})}$.

\subsection{Semi-supervised training scheme}
\label{sec:meanteacher}
The primary training methodology adopted in this work is based on the Mean Teacher framework \cite{tarvainen2017mean}. 
As illustrated in \cref{fig:main-arch}, we use two identical network architectures: a teacher-version, parameterized by ${\theta_\text{T}}$, and a student-version, described by $\theta_\text{S}$. In this setup, only the student is trained directly from data, $\mathbf{X}_L$ and $\mathbf{X}_U$. 
The teacher weights are updated after each training step using an exponentially moving average (EMA) of the student, ${\theta_T \leftarrow \alpha\theta_T + (1 - \alpha)\theta_S}$, where $\alpha$ is the keep rate.

We investigate two common forms of SSL schemes, pseudo-labelling and consistency regularization through feature similarity. The first follows the approach in \cite{berthelot2019remixmatch, sohn2020fixmatch, tarvainen2017mean, zou2019confidence}, where the student is trained using pseudo-labels (pseudo-maps for our setting) predicted by the teacher.
The teacher network receives a weakly augmented version of the input images $\mathbf{x}$ (and possibly a set of additional surrounding data samples, \cref{sec:fusion}), for which it predicts a pseudo-map ${\hat{\mathbf{y}}=\varphi(F_{\theta_{T}}(\mathbf{x}))}$. Here, $\varphi(\cdot)$ is used to decide/design suitable pseudo-labels based on the predicted class probabilities, \eg, thresholding and/or sharpening (see \cref{sec:sharp}).
The student receives a strongly augmented version of the same sample $\tilde{\mathbf{x}}$ (see \cref{sec:aug}) and is trained to make consistent class predictions with the teacher model through a pseudo-label loss ${\ell_\text{cls}(F_{\theta_S}(\tilde{\mathbf{x}}), \varphi(F_{\theta_T}(\mathbf{x})))}$ similar to $\ell_\text{sup}$.

The feature similarity supervision, on the other hand, enforces consistency between intermediate features for the teacher and student models. 
More formally, we train the student network to minimize a feature similarity loss, using \eg Mean Squared Error or Cosine Similarity, $\ell_\text{feat}(\mathbf{z}_{\theta_T}(\mathbf{x}), \mathbf{z}_{\theta_S}(\tilde{\mathbf{x}}))$. 
Here, $\mathbf{z}_{\theta_T}$ and $\mathbf{z}_{\theta_S}$ are intermediate features on the same level, \eg $\mathbf{z}_\text{bev}$ or $\tilde{\mathbf{z}}_\text{bev}$, using teacher and student models, respectively. 
In contrast to the pseudo-label training, feature similarity is insensitive to errors in teacher predictions as we avoid any label decisions. 
Instead, we enforce the representation of the same scene to be consistent in both models and is further discussed in \cref{sec:feat}.

In sum, the total training loss for the student is 
\begin{equation}
\begin{split}    
    \ell=\!\!\!\sum_{\mathbf{x}_i \in \mathbf{X}_L}\!\!\ell_{\text{sup}}(\mathbf{x}_i, \mathbf{y}_i)
    &+\!\!\!\sum_{\mathbf{x}_i \in \mathbf{X}_U}\!\!\omega_\text{cls}\ell_{\text{cls}}(F_{\theta_S}(\tilde{\mathbf{x}}_i), \varphi(F_{\theta_T}(\mathbf{x}_i))  \nonumber\\
    &+ \omega_\text{feat}\ell_{\text{feat}}(\mathbf{z}_{\theta_S}(\tilde{\mathbf{x}}_i), \mathbf{z}_{\theta_T}(\mathbf{x}_i))),
\end{split}    
\end{equation}
where $\omega_\text{cls}$ and $\omega_\text{feat}$ are hyperparameters controlling the relative importance of the respective loss.

\subsection{Augmentations}\label{sec:aug}
Strong augmentation techniques applied to the student’s input are essential for effective learning \cite{zhao2023augmentation}.
Therefore, we explore various data augmentation strategies that preserve the scene’s geometry, as discussed in \cref{sec:related-ssl}.
Specifically, we employ photometric augmentations, CamDrop \cite{ishikawa2024pct}, CutOut \cite{devries2017improved}, and BEVDrop \cite{yang2023revisiting}.
\emph{Photometric augmentations} introduce random variations in appearance by adjusting brightness, contrast, saturation, hue, and color conversion, as well as performing random channel swaps. 
\emph{CamDrop} augmentation simulates missing camera views by completely removing the input from one or more cameras. The dropped camera’s field of view is also masked from the loss.
\emph{CutOut} randomly masks out rectangular regions in the input images while
\emph{BEVDrop} operates on the BEV feature representation randomly setting certain grid cells to zero.

\subsection{Sharpening and thresholding}
\label{sec:sharp}
Predictions from the teacher are often refined through thresholding and sharpening before being used as pseudo-labels for the student. 
Thresholding keeps only predictions above some confidence level, whereas sharpening scales the predicted distribution to alter the confidence of the pseudo-label. 
For thresholding, we adapt FixMatch \cite{sohn2020fixmatch}, \ie, only predictions exceeding a fixed confidence threshold contribute to the loss. 
For sharpening, we explore two techniques: 
(1) \emph{hard} pseudo-labeling where teacher predictions are reduced to its argmax \cite{sohn2020fixmatch, lee2013pseudo}, and (2) temperature sharpening, where the predicted logits are rescaled through division by a temperature parameter (denoted $T$ in \cite{guo2017calibration, berthelot2019mixmatch}) before applying sigmoid to get a \emph{soft} pseudo-label. 
More formally, splitting up the classification head,
$\mathbf{p}_\text{cls} = f^\text{cls}_\theta (\mathbf{\tilde{z}}_{\text{bev}}) = \sigma(f^\text{logits}_\theta(\tilde{\mathbf{z}}_{\text{bev})})$, we have the intermediate logits $\mathbf{\hat{z}}$.
Sharpening using the hyperparameter $T$ is then defined as $\mathbf{\hat{z}}_{sharp} = \mathbf{\hat{z}} / T$.
For $ 0 < T < 1$ the certainty is artificially increased, \ie $\mathbf{\hat{z}}_{sharp} > \mathbf{\hat{z}}$. Conversely, $T > 1$ decreases the pseudo-label certainty.

\subsection{Feature Similarity}\label{sec:feat}
To further enhance consistency between the teacher and student networks, we impose feature similarity constraints at the BEV feature level.
A key consideration in this approach is selecting both the loss function and the feature level at which to apply the constraint.
We explore two candidate feature levels for enforcing similarity:
\emph{Early} BEV representation ($\mathbf{z}_{\text{bev}}$) – immediately after the lifting process, capturing the initial transformation of multi-view image features into the BEV space.
\emph{Late} BEV representation ($\tilde{\mathbf{z}}_{\text{bev}}$) – just before the final decoding step, where the network has refined its scene understanding.
By constraining intermediate features to be consistent, we encourage the student model to align with the teacher’s representations without relying on explicit pseudo-labels.
However, excessive weighting of the feature similarity loss may destabilize training, potentially leading to trivial solutions where the student merely replicates the teacher’s features instead of learning meaningful representations.

\subsection{Teacher Fusion} \label{sec:fusion}
Unlike BEV segmentation, which includes dynamic classes, all categories in online mapping are stationary.
This key distinction, combined with the fact that sensor data is collected in sequential frames with known ego vehicle motion, enables novel strategies not commonly explored in other domains.
We propose to leverage this by fusing pseudo-labels from the teacher across multiple nearby frames.

As illustrated in \cref{fig:main-arch}, predictions from both past and future frames can be transformed into the current coordinate frame using the known ego motion.
By selecting only the most confident teacher predictions from any of these frames, we provide the student with a more complete and accurate map representation.
This approach is particularly beneficial for improving predictions in regions far from the ego vehicle, where accuracy tends to degrade \cite{dong2023superfusion}.
We also explore feature fusion by averaging overlapping BEV feature representations ($\tilde{\mathbf{z}}_{\text{bev}}$) across multiple frames.

To ensure that additional frames contribute meaningful new information, we randomly sample them at varying distances relative to the current frame.
This strategy mitigates issues with strictly time-based selection, such as:
ego vehicle being stationary, which provides no new information, and the vehicle moving at high speeds, leading to minimal spatial overlap between frames.
\section{Experiments}
\label{sec:experiments}
\begin{table}[t]
    \centering
    \small
    \renewcommand{\arraystretch}{0.7}
    \setlength{\tabcolsep}{3pt}   
    \caption{Performance (mIoU) of various augmentations. Using \colorbox{YellowGreen!25}{ multiple augmentations} yields strong results, significantly outperforming \colorbox{Gray!25}{no augmentations}. We also note that \colorbox{Apricot!25}{CutOut \cite{devries2017improved}} works better than \colorbox{CornflowerBlue!25}{CamDrop \cite{ishikawa2024pct}} for online mapping.}
    \begin{NiceTabular}{cccccc}
    \CodeBefore
      \rowcolor{Grey!25}{3}
      \rowcolor{CornflowerBlue!25}{5}
      \rowcolor{Apricot!25}{6}
      \rowcolor{YellowGreen!25}{7}
    \Body
    \toprule
    \multicolumn{4}{c}{Augmentation} &  \multicolumn{2}{c}{Label utilisation} \\
    Photom. & CamDrop & CutOut & BEVDrop & $2.5\%$ & $10\%$\\ \midrule
     & & & &$12.1^{\pm0.6}$ & $21.3^{\pm0.2}$\\
    \cmark & & & & $13.3^{\pm0.6}$ & $32.0^{\pm0.1}$\\
     \cmark & \cmark & & & $21.9^{\pm1.0}$ & $31.9^{\pm0.1}$\\
     \cmark & & \cmark & & $24.4^{\pm1.2}$ & $33.9^{\pm0.2}$\\
    \cmark & & \cmark & \cmark & $\textbf{25.5}^{\pm0.7}$ & $\textbf{34.3}^{\pm0.1}$ \\
    \bottomrule
     \vspace{-15pt}
    \end{NiceTabular}
    \label{tab:augmentations}
\end{table}
%
\begin{figure}[t]
    \centering
    \includegraphics[width=\linewidth, trim={0mm, 13mm, 0mm, 0mm}, clip]{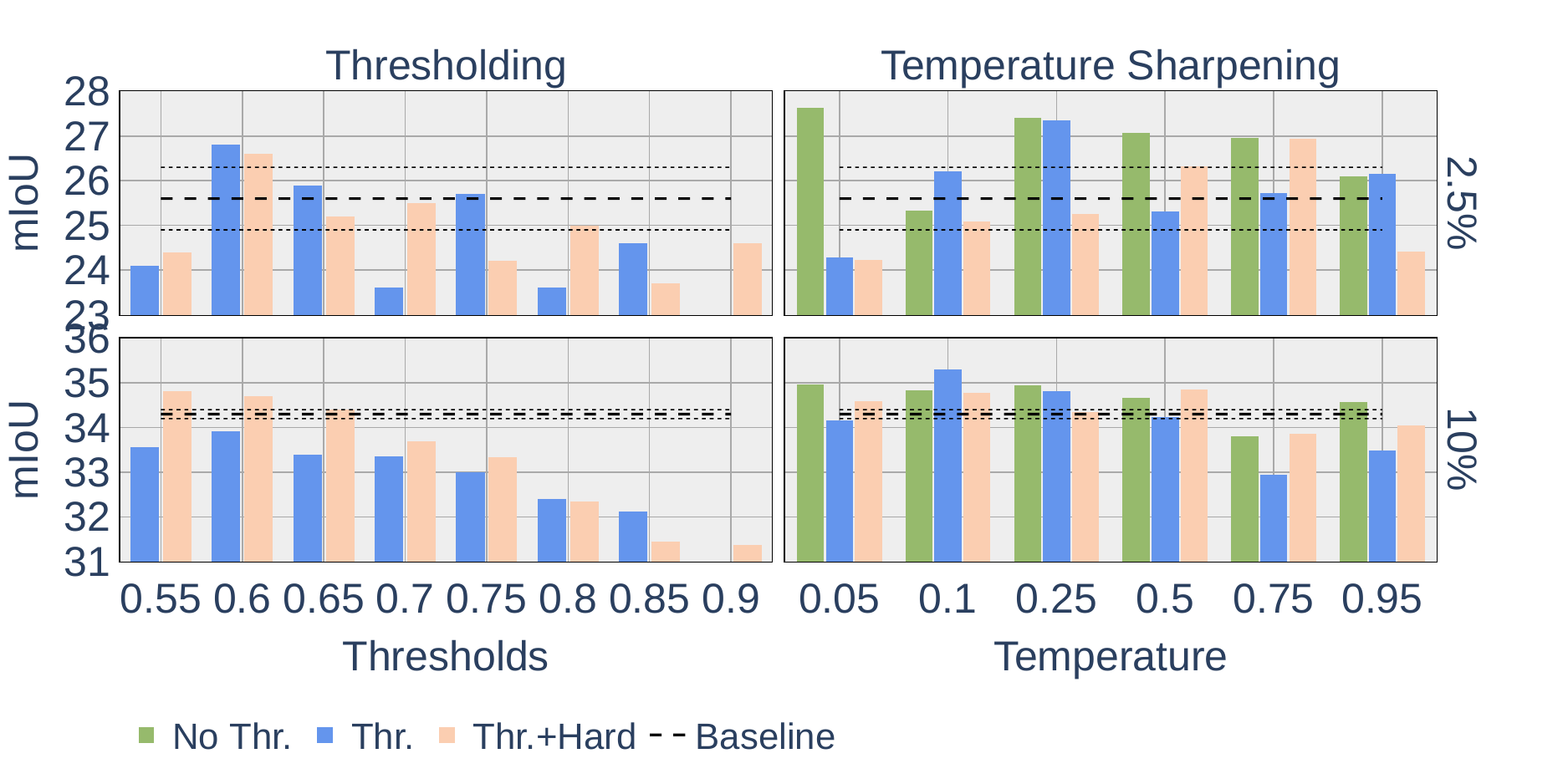}
    \caption{
    \colorbox{YellowGreen!25}{No thresholding}, soft \colorbox{CornflowerBlue!25}{MixMatch \cite{berthelot2019mixmatch}}, and hard \colorbox{Apricot!25}{FixMatch\cite{sohn2020fixmatch}} thresholding utilising $2.5\%$ and $10\%$ of the labels. 
    \protect\dashedbox{\emph{Baseline}}, mean and std., use neither thresholding nor sharpening.
     }
     \vspace{-15pt}
    \label{fig:threshold-sharpening}
\end{figure}
To explore the effectiveness of SSL for online mapping, we describe the experimental setup in \cref{sec:experimental-setup} and conduct comprehensive ablation studies in \cref{sec:exploration} to evaluate the contribution of each component. 
Building on these results, we further evaluate the performance in \cref{sec:exploitation}.

\subsection{Experimental Setup}
\label{sec:experimental-setup}
We employ a ResNet50 \cite{he2016deep} as image encoder $f^\text{enc}_\theta$, LSS \cite{philion2020lift} for lifting features to BEV $f^\text{lift}_\theta$, and a lightweight convolutional bottleneck decoder inspired by \cite{simplebev} as $f^\text{dec}_\theta$. 
We perform multi-label prediction for pedestrian crossings, lane dividers, and road boundaries and use Focal Loss \cite{lin2017focal} as the supervised loss $\ell_\text{sup}$ on the labelled dataset and $\ell_\text{cls}$ for the unlabelled data. 
The region of interest follows \cite{liu2024mgmap} spanning $\pm45$m longitudinally and $\pm15$m laterally. Each cell has a side length of $0.3$m.
For brevity, we report the mean Intersection over Union (mIoU) across these three classes.
To account for the stochasticity in the results from individual runs, we either report the mean and standard deviation of the mIoU for three runs with the same hyperparameters or multiple nearby values of the hyperparameter.
We choose to follow \cite{gao2022s2g2} and use unlabelled weight $\omega_{\text{cls}}=1$.
We follow \cite{ishikawa2024pct} and linearly ramp-up $\omega_{\text{cls}}$, and when applicable $\omega_{\text{feat}}$, over the first third of the training.
Unless stated otherwise, no thresholding or sharpening is used in the experiments.
For the CutOut augmentation, we remove $25\%$ of the image, and for BEV feature dropout (BEVDrop) we follow \cite{ishikawa2024pct, yang2023revisiting} and drop $50\%$ of the features.
The models, unless stated otherwise, are trained and evaluated using the Argoverse 2 (AV2) dataset \cite{argoverse2_2021} with the geographically disjoint splits proposed by \cite{lilja2024localization} using surround-view camera images as input and use the provided HD maps as labels.
We generally experiment using two utilisation rates of labelled data, namely $2.5\%$ and $10\%$, corresponding to $18$ and $70$ labelled sequences respectively. 

\subsection{Exploration}
\label{sec:exploration}
\begin{table}[t]
    \centering
    \renewcommand{\arraystretch}{0.7}
    \caption{mIoU for sharpening methods using threshold $0.6$. 
     Neither soft \colorbox{CornflowerBlue!25}{MixMatch \cite{berthelot2019mixmatch}} nor hard thresholding \colorbox{Apricot!25}{FixMatch\cite{sohn2020fixmatch}} show significantly impact over using \colorbox{YellowGreen!25}{no thresholding}
     }
    \begin{NiceTabular}{cccc}
    \CodeBefore
      \rowcolor{YellowGreen!25}{3}
      \rowcolor{CornflowerBlue!25}{4}
      \rowcolor{Apricot!25}{5}
    \Body
    \toprule
    \multicolumn{2}{c}{Technique} & \multicolumn{2}{c}{Label utilisation} \\ 
    Thr.   & Hard  & $2.5\%$ & $10\%$ \\ \midrule
          &        & $25.5^{\pm0.7}$ & $34.3^{\pm0.1}$ \\
    \cmark &        & $25.1^{\pm1.3}$ & $33.9^{\pm0.1}$ \\
    \cmark & \cmark & $25.4^{\pm0.8}$ & $34.6^{\pm0.1}$ \\ 
    \bottomrule
    \vspace{-20pt}
    \end{NiceTabular}
    \label{tab:threshold-sharpening}
\end{table}
We explore different SSL approaches outlined in \cref{sec:method} to show their effectiveness for segmentation-based online mapping.

\parsection{Augmentations}
We evaluate various strong augmentation strategies for the student model, as discussed in \cref{sec:aug}.
The results in \cref{tab:augmentations} demonstrate that increasing the difficulty of the task for the student model is crucial also for online mapping.
With $10\%$ label utilisation, Photometric augmentation alone is sufficient for significant model improvement. However, with $2.5\%$ label utilisation, additional augmentations are required.
For both label utilisation ratios, applying Photometric, CutOut, and BEVDrop yields the best performance, and we adopt this configuration as the \emph{Baseline} in further experiments. 
Specifically, we observe a significant performance improvement, with mIoU increasing from $12.1$ to $25.5$ for the $2.5\%$ label utilisation and $21.3$ to $34.3$ for the $10\%$ label utilization.
Furthermore, contrary to the findings in \cite{ishikawa2024pct}, we find that removing parts of the images, \ie CutOut, yields better performance than dropping an entire camera and masking its field of view in the loss, as done with CamDrop.

\parsection{Sharpening and Thresholding}
The effects of sharpening and thresholding are illustrated in \cref{fig:threshold-sharpening}.
As shown to the left, supervising with the most confident pseudo-labels does not yield significant performance improvements.
However, using a low confidence threshold provides a slight boost in performance, and we set the threshold to 0.6 for subsequent experiments.
The right plot shows the impact of sharpening. 
While sharpening can lead to minor improvements, the gains are marginal and introduce an additional hyperparameter, making it less appealing for our setting.
\cref{tab:threshold-sharpening} presents results across multiple runs, comparing hard and soft label supervision.
Overall, these techniques have limited impact on online mapping, contrasting findings in other domains where sharpening has been highly effective—typically when used alongside high thresholds \cite{sohn2020fixmatch}.

\begin{figure}
    \centering
    \includegraphics[width=\linewidth, trim={0mm, 0mm, 0mm, 17mm}, clip]{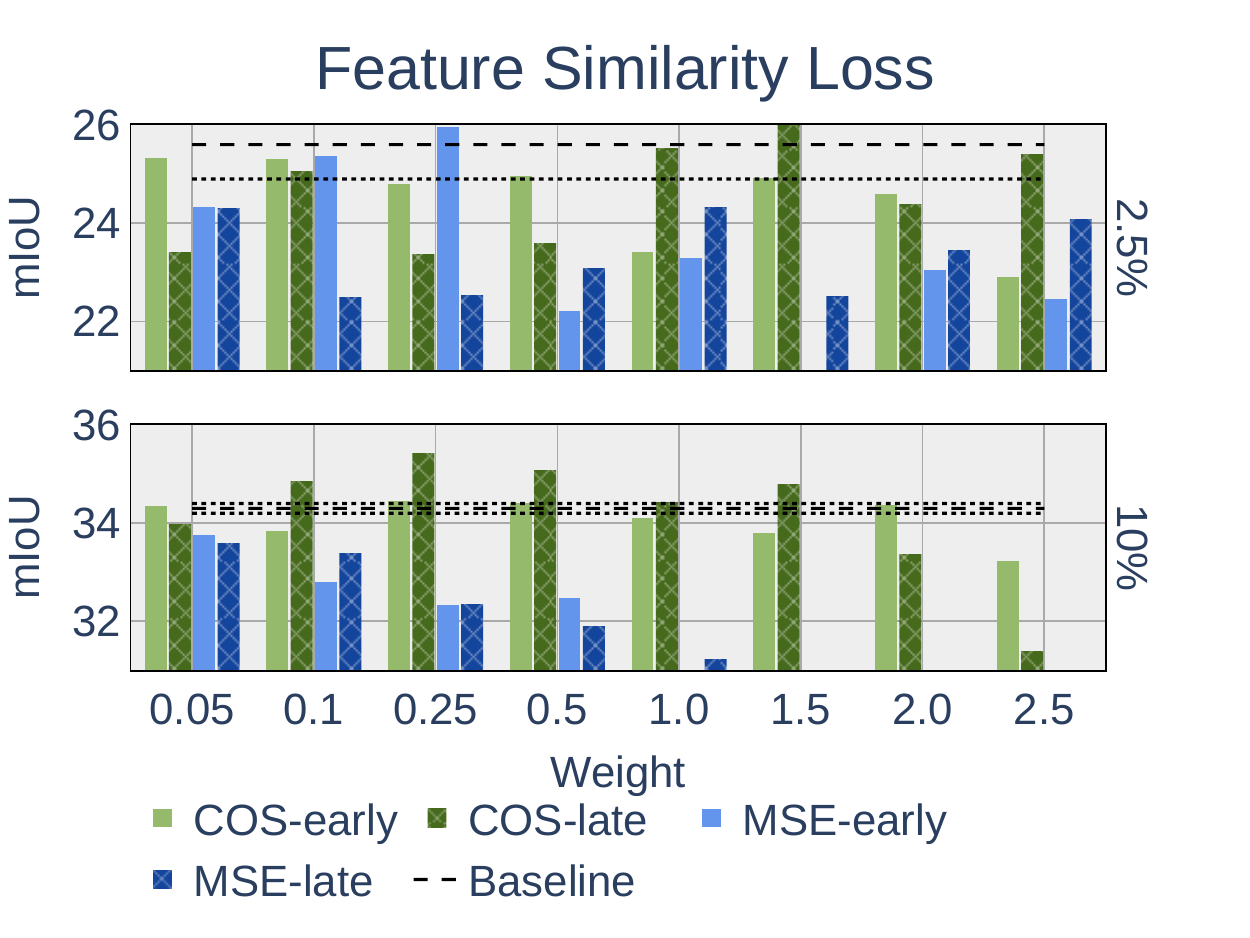}
    \caption{Feature Similarity using Mean Square Error following \cite{zhu2024semibevseg} and Cosine Similarity following \cite{wallin2022doublematch} varying $\omega_{\text{feat}}$. 
    \protect\dashedbox{\emph{Baseline}} refers to no feature similarity while \emph{early} and \emph{late} refers to using BEV features before or after BEV processing $f^{\text{dec}}_\theta$ in \cref{fig:main-arch}.}
    \label{fig:featsim}
    \vspace{-10pt} 
\end{figure}

\parsection{Feature Consistency}
We compare the performance using Mean Squared Error (MSE) and Cosine Similarity (COS) losses with $\omega_{\text{cls}} = 1$ and varying feature loss weight $\omega_{\text{feat}}$. 
In \cref{fig:featsim} the results are displayed using features from before or after the BEV decoder $f^{\text{dec}}_\theta$ as inputs to $\ell_\text{feat}$.
When utilizing $2.5\%$ of the labels the results are noisy, and no clear benefit from the additional supervision is observed across the setups.
However, for the $10\%$ label utilisation, \emph{late} COS consistency provides a slight performance gain and demonstrates greater stability than the MSE loss. 
While prior BEV segmentation works opted to use MSE \cite{zhu2024semibevseg, gao2022s2g2}, we believe COS to be the better choice for our setting.

\begin{figure}
    \centering
\includegraphics[width=\linewidth, trim={0mm, 0mm, 0mm, 18mm}, clip]{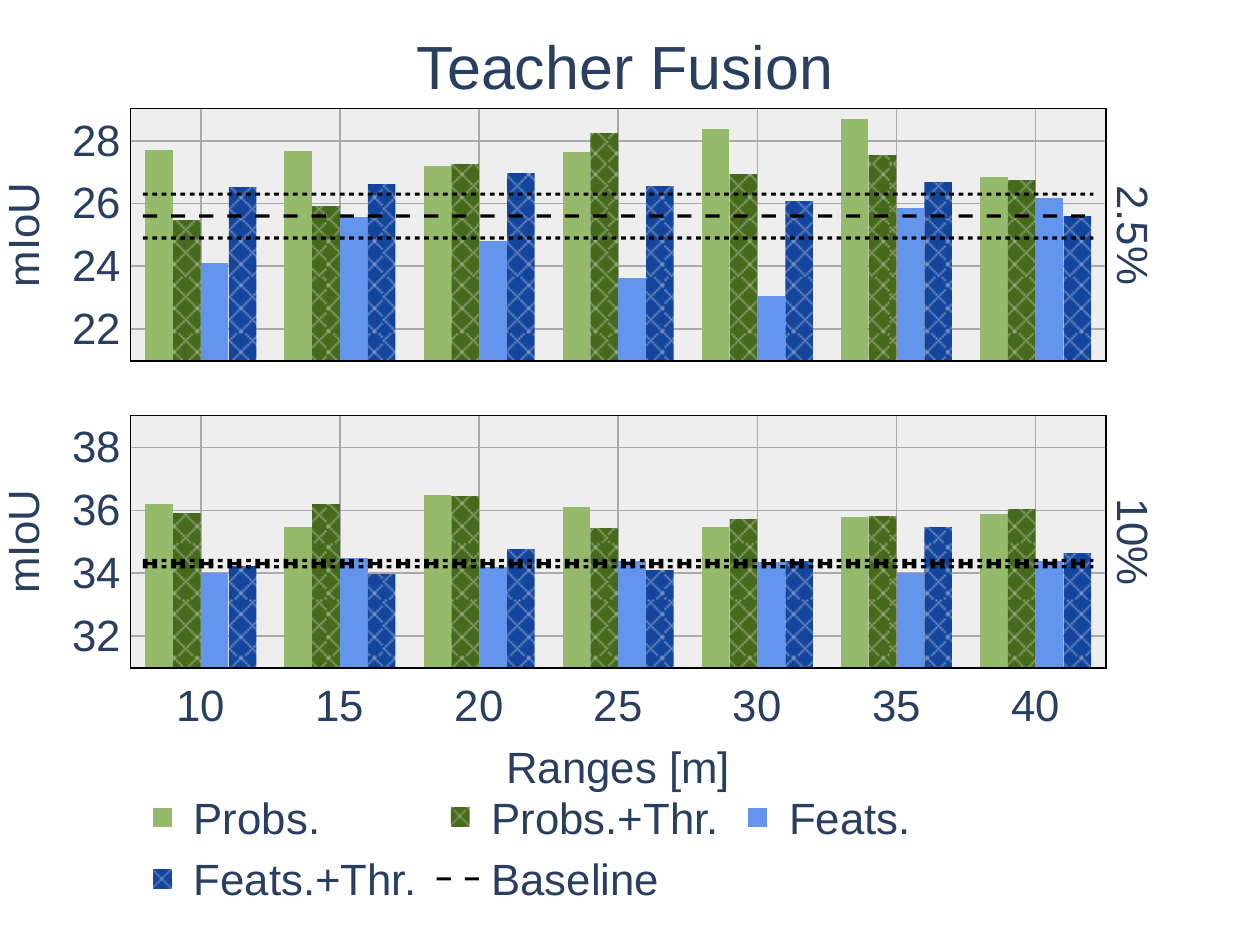}
    \caption{Teacher Fusion across varying maximum ranges from the current sample. 
    Fusing the probabilities yields consistent improvements over the \protect\dashedbox{\emph{Baseline}} without any fusion.
    }
    \label{fig:fusion-dist}
    \vspace{-15pt}
\end{figure}

\parsection{Teacher Fusion}
To enhance pseudo-labels, we propose to fuse teacher predictions from multiple samples. 
We use two additional samples, as we empirically found, see \supmat~\cref{supmat:exact-numbers}, that adding more samples does not provide further benefits but makes training slower due to the additional inference costs. 
Therefore, we focus on examining the influence of the maximum distance to these additional samples.
The results in \cref{fig:fusion-dist} show that even our simple fusion of predicted probabilities generally improves performance, both with and without thresholding.
The fusion of BEV features, on the other hand, shows no significant benefit. 
For $2.5\%$ label utilisation we observe that increasing the range to above $20$m is required to improve performance, whereas for $10\%$ the maximum range is not as important. 
In further experiments, we will use $30$m as it works well for both utilisation ratios.
While the simple probability fusion strategy yields strong results, exploring more advanced fusion techniques is a promising avenue for future research.

\parsection{Combining Components}
\label{sec:ablation}
Building on the above results, we further explore how these components interact in SSL for online mapping. 
Starting with the teacher-student setup as a \emph{Core} we incrementally add components and evaluate their impact on performance.
Here, we add strong augmentations to the student, \emph{+Augs}, with Photometric, CutOut, and BEVDrop. 
Teacher fusion, \emph{+Fusion}, is used with a maximum range of $30$m. 
The Cosine feature similarity loss is incorporated under \emph{+Featsim} with a weight of $w_{\text{feat}}=0.25$.
Additionally, a threshold of $0.6$ is used without any temperature sharpening in \emph{+Thr}, and hard pseudo-labeling is further added and denoted with \emph{+Hard}.

The results in \cref{tab:ablation} demonstrate that combining components leads to further increased performance. 
Adding strong augmentations to the student yields a significant performance boost, as does further incorporating teacher fusion. 
Together these contribute to mIoU gains of $16.1$ and $15.1$ for $2.5\%$ and $10\%$ respectively.
When combined with Fusion, feature similarity slightly improves performance for $2.5\%$ labelled data, but has no impact for $10\%$. 
Finally, applying thresholding yields a modest improvement for both ratios, while using the hard pseudo-labels does not contribute further.
These results underscore the importance of conducting ablation studies across different labelled data ratios, as the impact of various modules can vary depending on the amount of labelled data.

\begin{table}[t]
    \centering
    \small
    \renewcommand{\arraystretch}{0.8}
    \setlength{\tabcolsep}{1pt}    
    \caption{mIoU when incrementally adding SSL components.} 
    \begin{tabular}{ccccccc}
    \toprule
    Lbl. Util. & Core & +Augs & +Fusion & +Featsim & +Thr. & +Hard \\ \midrule
    $2.5\%$ & $12.1^{\pm0.6}$ & $25.5^{\pm0.7}$ & $28.2^{\pm0.7}$ & $29.0^{\pm0.6}$ & $29.4^{\pm0.3}$ & $29.3^{\pm1.1}$ \\
    $10\%$ & $21.3^{\pm0.2}$ & $34.3^{\pm0.1}$ & $36.4^{\pm0.2}$ & $36.4^{\pm0.5}$ & $36.8^{\pm0.2}$ & $36.7^{\pm0.3}$ \\
    \bottomrule
    \vspace{-10pt}
    \end{tabular}
    \label{tab:ablation}
\end{table}

\subsection{Putting SSL to the Test}
\label{sec:exploitation}
Building on the design choices made above, we aim to further verify the strengths of our proposed SSL scheme. 
Using the parameters from \cref{sec:ablation}, our proposed method includes strong augmentations, teacher fusion, cosine feature similarity, and thresholding. 
As before, we report the mean and standard deviation; however, in this case, we select the best-performing model on the validation set and evaluate it on the test set.
The performance is compared by utilizing both unlabelled and labelled data with the \colorbox{CornflowerBlue!25}{SSL-scheme} versus using \emph{only} the labelled data in a \colorbox{BrickRed!25}{supervised} setting. 
First, we compare our SSL scheme against swapping the backbone for a vision foundation model. 
Next, we evaluate the robustness of our method across different online mapping architectures. 
We then examine its scalability with varying amounts of labelled data and demonstrate the method's effectiveness for domain adaptation.

\parsection{Comparing with pre-training}
\label{sec:compare-w-dino}
We substitute the ImageNet pre-trained ResNet50 with DINOv2 \cite{oquab2024dinov2} vision foundation model, which leverages pre-training on a substantially larger dataset. 
Although DINOv2's size makes it impractical for vehicular deployment, it demonstrates performance improvements in \cref{tab:dino-backbone}. 
Nevertheless, a significant performance gap persists when comparing DINOv2 without SSL to our SSL approach. 
Notably, integrating our SSL scheme with the DINOv2 backbone eliminates this gap, underscoring the effectiveness of our approach.

\begin{table}[]
    \centering
    \small
    \renewcommand{\arraystretch}{0.5}
    \caption{mIoU when when using DINOv2 backbone adding SSL.} 
    \begin{tabular}{lccc}
    \toprule
    \multirow{2}{*}{Backbone} & \multirow{2}{*}{SSL} & \multicolumn{2}{c}{Label utilisation} \\ 
               &         & $2.5\%$ & $10\%$      \\ \midrule
    ResNet50   &         & $7.5^{\pm 0.9}$       & $10.6^{\pm 0.2}$ \\ 
    DINOv2     &         & $11.3^{\pm 0.6}$      & $17.3^{\pm 0.4}$\\ \midrule
    ResNet50   & \cmark  & $29.4^{\pm 0.3}$      & $36.8^{\pm 0.2}$ \\ 
    DINOv2     & \cmark  & $28.7^{\pm 0.9}$      & $33.2^{\pm 0.2}$\\
    \bottomrule
    \vspace{-20pt}
    \end{tabular}
    \label{tab:dino-backbone}
\end{table}

\begin{figure}[]
    \centering
    \includegraphics[width=0.9\linewidth, trim={0mm, 7mm, 0mm, 0mm}, clip]{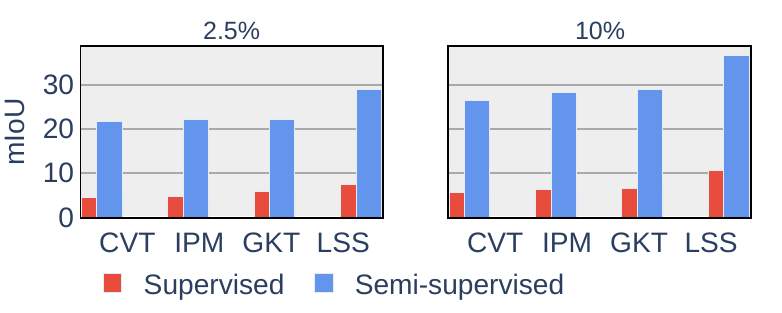}
    \caption{Multiple SOTA online mapping methods benefit from the \colorbox{CornflowerBlue!25}{SSL-scheme} compared to the \colorbox{BrickRed!25}{supervised} training.}
    \label{fig:performance-model-archs}
     \vspace{-10pt}
\end{figure}

\parsection{Robustness to model architecture}
\label{sec:robustness-archs}
We apply our proposed SSL setup to state-of-the-art online mapping models. 
As shown in \cref{fig:performance-model-archs}, various architectures—including CVT \cite{zhou2022crossview}, IPM \cite{simplebev}, GKT \cite{chen2022efficient}, and LSS \cite{philion2020lift} -- all benefit significantly from the SSL scheme. LSS achieves the highest overall performance, with a $3.9$x and $3.5$x boost when $2.5\%$ and $10\%$ of the data have labels, respectively. 
While \eg CVT performs slightly worse overall, it sees even greater relative improvements, with performance increasing by $4.8$x and $4.6$x under the same label utilizations.

\parsection{Scalability}
\label{sec:varying-label-util}
\begin{figure}[t]
    \centering
    \begin{subfigure}[t]{\linewidth}
    \includegraphics[width=\linewidth, trim={0mm, 17mm, 0mm, 20mm}, clip]{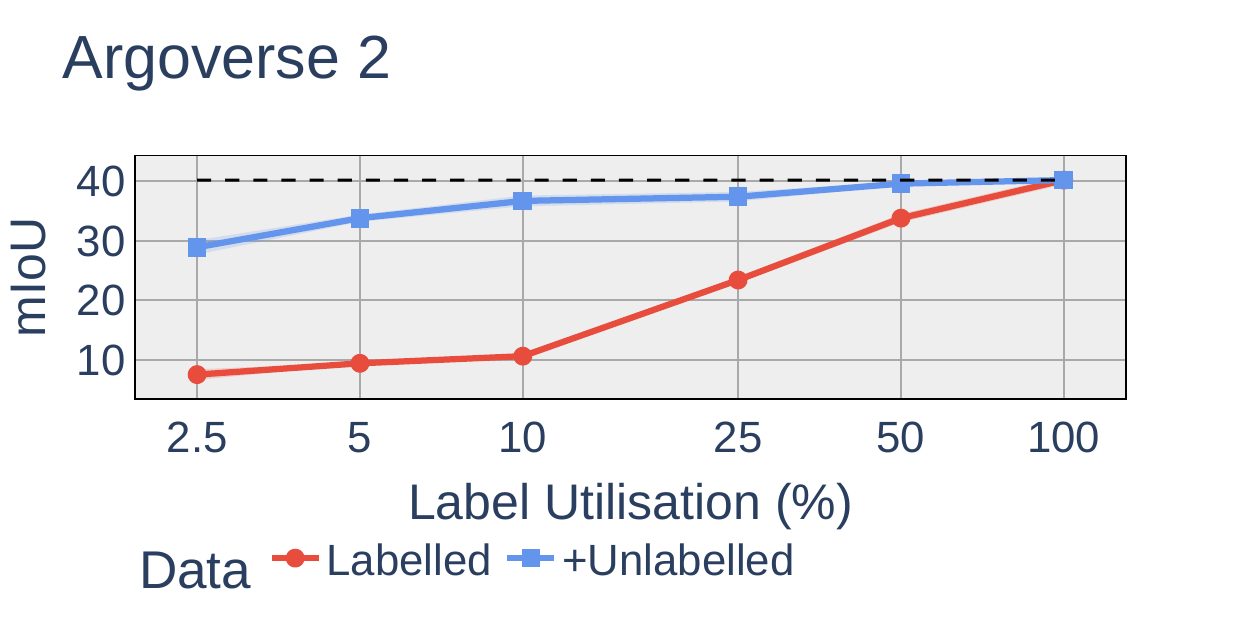}
    \caption{Argoverse 2}
    \label{fig:main-results-av2}
\end{subfigure}
\begin{subfigure}[t]{\linewidth}
    \centering
    \includegraphics[width=\linewidth, trim={0mm, 17mm, 0mm, 20mm}, clip]{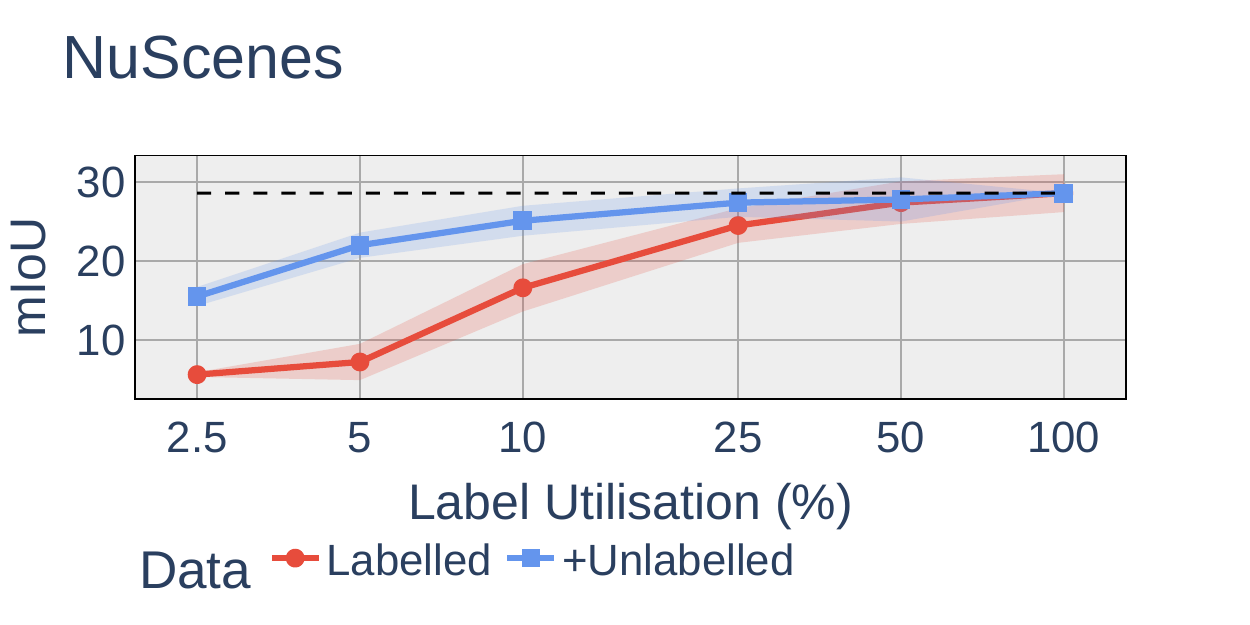}
    \caption{nuScenes}
    \label{fig:main-results-nusc}
    \end{subfigure}
    \caption{Test performance (mIoU) across utilising various amount of the labelled nuScenes data.
    All data is used as unlabelled for all runs. 
    Training with unlabelled data using the \colorbox{CornflowerBlue!25}{SSL-scheme} consistently improves performance compared to only using \colorbox{BrickRed!25}{supervised} training across all label utilisations.  
     }
      \label{fig:main-results}
      \vspace{-10pt}
\end{figure}
\begin{figure}[]
    \centering
    \includegraphics[width=\linewidth]{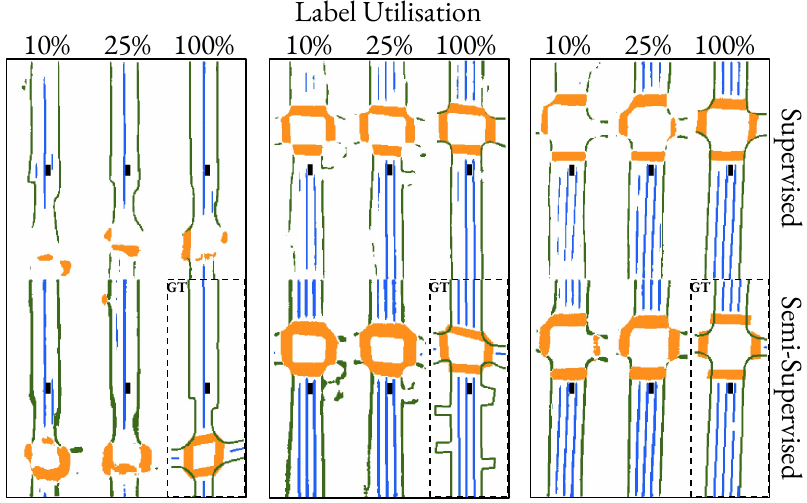}
    \caption{Qualitative results with varying amounts of labelled data.
Predictions are shown for models trained with $10\%$, $25\%$, and $100\%$ of the labelled data, using both standard supervised training and our proposed SSL approach. 
}
    \label{fig:qualitative-examples}
     \vspace{-10pt}
\end{figure}
\begin{figure}[t]
        \centering
    \includegraphics[width=\linewidth, trim={5mm, 12mm, 21mm, 8mm}, clip]{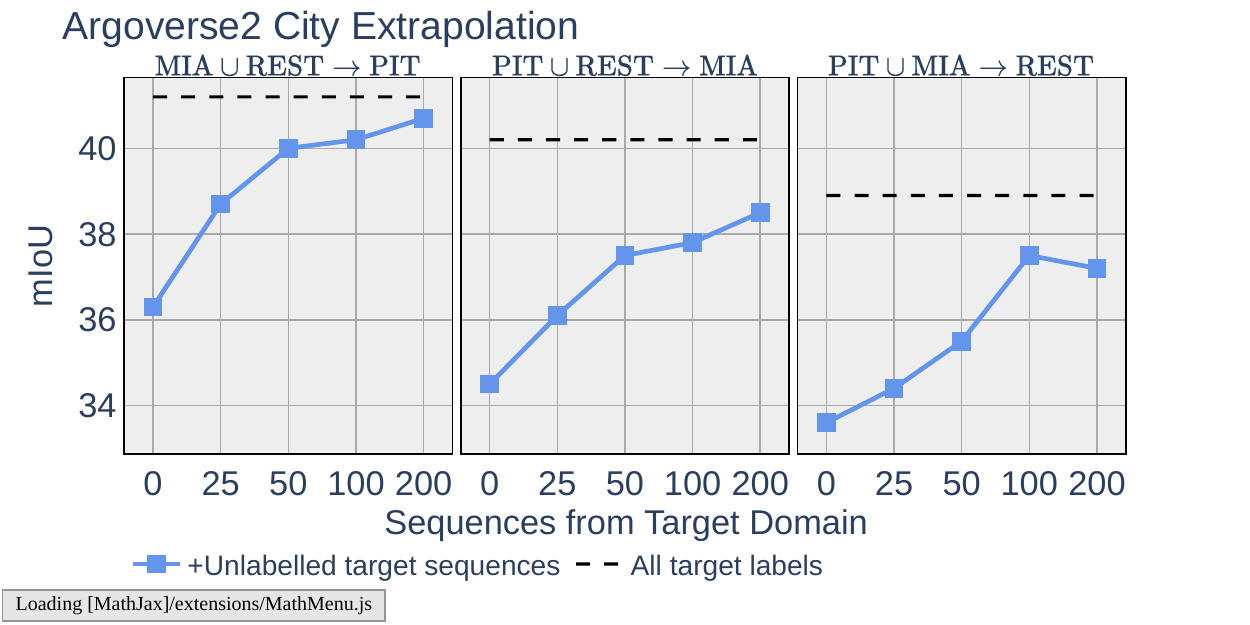}
        \caption{Adapting to new cities for AV2. 
       Adding unlabelled sequences with the \colorbox{CornflowerBlue!25}{SSL-scheme} from the target domain enhances performance in the target city. 
       When adapting to Pittsburgh, the gap is almost completely closed to using \protect\dashedbox{all target labels}.
        }
        \label{fig:citywise-av2}
         \vspace{-15pt}
    \end{figure}
We conduct experiments on AV2 \cite{argoverse2_2021} and nuScenes \cite{nuscenes_cvpr_2020}, with the same hyperparameters except for learning rate and weight decay. 
Our experiments range from using the labels from the full dataset, corresponding to $700$ sequences, down to just $18$ sequences at $2.5\%$ utilisation. 
\Cref{fig:main-results} shows how our proposed SSL method and its supervised counterpart scale with varying amounts of labelled data.
For the supervised models, performance decreases rapidly as the amount of (labelled) data is reduced. 
In the most label-scarce setting, having access to $2.5\%$ of the labels, models trained solely on labelled sequences perform poorly, with mIoU dropping to $7.5$ for AV2 and $5.6$ for nuScenes. 
Incorporating unlabelled data with our proposed approach yields substantial performance boosts across all labelled data utilisation ratios. 
Leveraging unlabelled data with the SSL-scheme dramatically improves performance to $28.9$ and $15.5$ mIoU for the respective datasets. 

If $10\%$ of the data has labels, our method achieves a $3.5$x performance boost compared to only using the labelled data. 
This narrows the gap to a fully supervised model that uses all labels, to just $3.5$ mIoU on AV2.
For nuScenes, the performance boost is $1.5$x, and the remaining gap to when all labels have been used is reduced to $3.5$ mIoU.

To provide visual context for these quantitative improvements, \cref{fig:qualitative-examples} shows predictions from models trained only on labelled data against those trained with both labelled and unlabelled data.
The predictions are noticeably more stable and reliable when using our semi-supervised method, aligning with the quantitative improvements reported above.
 
\begin{figure}[t]
        \centering
        \vspace{0.5mm}
\includegraphics[width=1\linewidth, trim={0mm, 27mm, 0mm, 6mm}, clip]{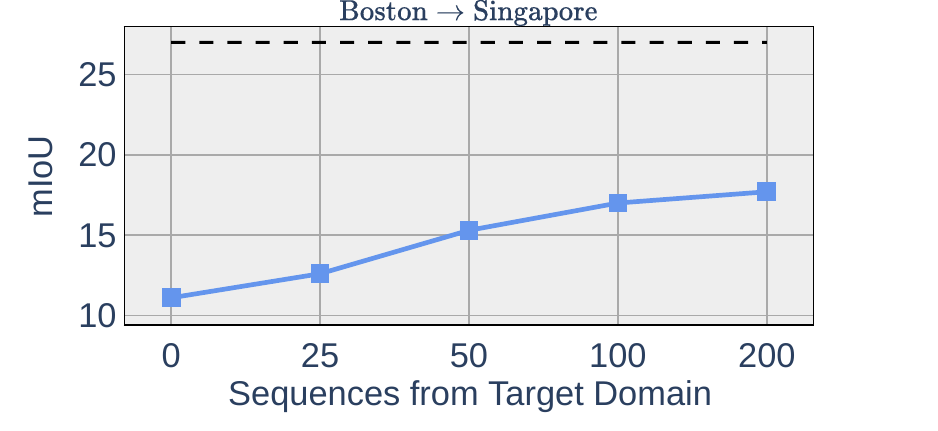}
        \caption{City adaptation from Boston to Singapore on nuScenes. 
        While performance naturally is lower than using \protect\dashedbox{all Singapore labels}, incorporating unlabelled sequences using the \colorbox{CornflowerBlue!25}{SSL-scheme} helps improve the model's performance when labelled data is only available from Boston.
        }
        \label{fig:citywise-nusc}
         \vspace{-15pt}
\end{figure}

\parsection{City Adaptation}
\label{sec:city-adaptation}
We assess our method’s ability to adapt to new cities using the AV2 Far Extrapolation splits from \cite{lilja2024localization} and the Boston-to-Singapore split from \cite{man2023dualcross} for nuScenes.
Our approach utilises all labelled data from the source cities while gradually incorporating unlabelled data from the target cities, ensuring geographic separation from the test set.
As shown in \cref{fig:citywise-av2} and \cref{fig:citywise-nusc} for AV2 and nuScenes, respectively, we find that even without labelled data from the target city, performance improves significantly by utilizing its unlabelled data.
The domain gap between Boston and Singapore in nuScenes is substantial, making adaptation more challenging than in AV2, where all data comes from US cities.
Notably, performance improves as more unlabelled sequences from the target domain are added, suggesting that additional unlabelled data could further enhance results.
For smaller domain gaps, such as adapting to Pittsburgh in AV2, our method nearly matches the performance of a fully supervised model, with only a $0.5$ mIoU difference.
For larger domain gaps, such as Boston to Singapore, our method boosts mIoU from $11.1$ to $17.0$ when using $100$ unlabelled sequences from Singapore.
However, a substantial gap remains in cases of large domain shifts, posing a key challenge for future research.

\section{Conclusion}
\label{sec:conclusion}
This paper draws inspiration from SSL techniques in other domains, demonstrating their effectiveness for segmentation-based online mapping.
We show that applying photometric, CutOut, and BEVDrop augmentations in a teacher-student setup already performs well.
When combined with our proposed multi-step fusion, and feature similarity constraints we show an additional performance boost.

Using the proposed SSL-scheme, we show that incorporating unlabelled data substantially improves performance on both AV2 and nuScenes compared to relying solely on labelled data. 
For example, if $10\%$ of the data has labels, the SSL-approach to leverage unlabelled data achieves a $3.5$x performance boost compared to only using the labelled data, narrowing the gap to a fully supervised model that uses all labels to just $3.5$ mIoU.
We also show strong generalization to unseen cities and believe that additional unlabelled data could further enhance the domain adaptation results. 
Still, when adapting to Pittsburgh (AV2), incorporating purely unlabelled target-domain data reduces the performance gap from $5$ to just $0.5$ mIoU.

In summary, efficiently leveraging both labelled and unlabelled data is crucial for advancing online mapping. 
While this work does not explore all existing SSL methods, it provides a foundation for understanding key SSL methods in this context and highlights their potential.
We hope this study inspires further research in teacher fusion strategies, self-supervised pre-text tasks, and broader semi-supervised learning approaches to enhance online mapping performance further.

\newpage
\section*{Acknowledgements}
We want to express our sincere gratitude to Adam Tonderski, Georg Hess, William Ljungbergh, Carl Lindström, Ji Lan, and Erik Stenborg for their continuous support and invaluable discussions.
This work was partially supported by the Wallenberg AI, Autonomous Systems and Software Program (WASP) funded by the Knut and Alice Wallenberg Foundation. Computational resources were provided by the National Academic Infrastructure for Supercomputing in Sweden (NAISS) at \href{https://www.nsc.liu.se/}{NSC Berzelius} and \href{https://www.c3se.chalmers.se/about/Alvis/}{C3SE Alvis} partially funded by the Swedish Research Council through grant agreement no. 2022-06725.

{
    \small
    \bibliographystyle{ieeenat_fullname}
    \bibliography{main}
}

\clearpage
\setcounter{page}{1}
\maketitlesupplementary

\section{Exact numbers} 
\label{supmat:exact-numbers}
For completeness, we also show the exact performance numbers for the results shown in Figures in \cref{sec:experiments}. For thresholds and temperatures, the numbers are displayed in \cref{tab:thresholds-2p5} and \cref{tab:temperatures-2p5} for $2.5\%$ of the labels, respectively. For $10\%$, we refer to \cref{tab:thresholds-10p0} and \cref{tab:temperatures-10p0}.
In addition to these, the performance for feature similarity loss with different weights is shown in \cref{tab:featuresim-2p5} and \cref{tab:featuresim-10p0} for $2.5\%$ and $10\%$ label utilisation, respectively.
The validation performance for teacher fusion over various ranges is provided in \cref{tab:teacherfusion-2p5} and \cref{tab:teacherfusion-10p0} for $2.5\%$ and $10\%$ label utilisation.
We also experiment with the amount of samples to use in addition to the center sample in \cref{tab:fusionframes}.
For the test performance across multiple label utilisations, the numbers for both Argoverse 2 (AV2) and nuScenes are shown in \cref{tab:alpha-runs}.
Finally, \cref{tab:domainadapt} provides the performance for city adaptation for both Argoverse 2 and nuScenes when varying the amount of unlabelled data.

\begin{table}[h!]
    \small
    \setlength{\tabcolsep}{2pt}   
    \centering
    \caption{Validation performance (mIoU) for different thresholds on Argoverse 2 $2.5\%$ of the labels utilised.} 
    \begin{tabular}{lrrrrrrrr}
    \toprule
    Thresholds & $  0.55$ & $  0.6$ & $  0.65$ & $  0.7$ & $  0.75$ & $  0.8$ & $  0.85$ & $  0.9$ \\
    \midrule
    Thr.      & $24.1$ & $26.8$ & $25.9$ & $23.6$ & $25.7$ & $23.6$ & $24.6$ & $22.3$ \\
    Thr.+Hard & $24.4$ & $26.6$ & $25.2$ & $25.5$ & $24.2$ & $25.0$ & $23.7$ & $24.6 $\\
    \bottomrule
    \end{tabular}
    \label{tab:thresholds-2p5}
\end{table}

\begin{table}[h!]
    \small
    \centering
    \caption{Validation performance (mIoU) for using sharpening with different temperatures on Argoverse 2 $2.5\%$ of the labels utilised.} 
\begin{tabular}{lrrrrrr}
\toprule
Temperature & $  0.05$ & $  0.1$ & $  0.25$ & $  0.5$ & $  0.75$ & $  0.95$ \\
\midrule
No Thr. & $  27.6$ & $  25.3$ & $  27.4$ & $  27.1$ & $  27.0$ & $  26.1 $\\
Thr. & $  24.3$ & $  26.2$ & $  27.4$ & $  25.3$ & $  25.7$ & $  26.2 $\\
Thr.+Hard & $  24.2$ & $  25.1$ & $  25.3$ & $  26.3$ & $  26.9$ & $  24.4 $\\
\bottomrule
\end{tabular}
    \label{tab:temperatures-2p5}
\end{table}

\begin{table}[h!]
    \small
    \setlength{\tabcolsep}{3pt}   
    \centering
    \caption{Validation performance (mIoU) for different thresholds on Argoverse 2 $10\%$ of the labels utilised.} 
\begin{tabular}{lrrrrrrrr}
\toprule
Thresholds & $  0.55$ & $  0.6$ & $  0.65$ & $  0.7$ & $  0.75$ & $  0.8$ & $  0.85$ & $  0.9 $\\
\midrule
Thr. & $  33.6$ & $  33.9$ & $  33.4$ & $  33.3$ & $  33.0$ & $  32.4$ & $  32.1$ & $  30.9 $\\
Thr.+Hard & $  34.8$ & $  34.7$ & $  34.4$ & $  33.7$ & $  33.3$ & $  32.3$ & $  31.4$ & $  31.4 $\\
\bottomrule
\end{tabular}
    \label{tab:thresholds-10p0}
\end{table}

\begin{table}[h!]
    \small
    \centering
    \caption{Validation performance (mIoU) for using sharpening with different temperatures on Argoverse 2 $10\%$ of the labels utilised.} 
\begin{tabular}{lrrrrrr}
\toprule
Temperature & $0.05$ & $0.1$ & $0.25$ & $0.5$ & $0.75$ & $0.95$ \\
\midrule
No Thr. & $  35.0$ & $  34.8$ & $  34.9$ & $  34.7$ & $  33.8$ & $  34.6 $\\
Thr. & $  34.2$ & $  35.3$ & $  34.8$ & $  34.2$ & $  32.9$ & $  33.5 $\\
Thr.+Hard & $  34.6$ & $  34.8$ & $  34.4$ & $  34.9$ & $  33.9$ & $  34.1 $\\
\bottomrule
\end{tabular}
    \label{tab:temperatures-10p0}
\end{table}

\begin{table}[h!]
    \small
    \centering
    \caption{Validation performance (mIoU) for feature similarity loss with different weights on Argoverse 2 $2.5\%$ of the labels utilised.} 
\begin{tabular}{lrrrrrrrr}
\toprule
Weight & $  0.05$ & $  0.1$ & $  0.25$ & $  0.5$ & $  1.0$ & $  1.5$ & $  2.0$ & $  2.5 $\\
\midrule
MSE-pre & $  24.3$ & $  25.4$ & $  26.0$ & $  22.2$ & $  23.3$ & $  20.0$ & $  23.1$ & $  22.5 $\\
MSE-post & $  24.3$ & $  22.5$ & $  22.5$ & $  23.1$ & $  24.3$ & $  22.5$ & $  23.4$ & $  24.1$ \\
COS-pre & $  25.3$ & $  25.3$ & $  24.8$ & $  25.0$ & $  23.4$ & $  24.9$ & $  24.6$ & $  22.9 $\\
COS-post & $  23.4$ & $  25.1$ & $  23.4$ & $  23.6$ & $  25.5$ & $  26.1$ & $  24.4$ & $  25.4$ \\
\bottomrule
\end{tabular}
    \label{tab:featuresim-2p5}
\end{table}

\begin{table}[h!]
    \small
    \centering
    \caption{Validation performance (mIoU) for feature similarity loss with different weights on Argoverse 2 $10\%$ of the labels utilised.} 
\begin{tabular}{lrrrrrrrr}
\toprule
Weight & $  0.05$ & $  0.1$ & $  0.25$ & $  0.5$ & $  1.0$ & $  1.5$ & $  2.0$ & $  2.5$ \\
\midrule
MSE-pre & $  33.7$ & $  32.8$ & $  32.3$ & $  32.5$ & $  32.4$ & $  30.0$ & $  28.8$ & $  24.1$ \\
MSE-post & $  33.6$ & $  33.4$ & $  32.3$ & $  31.9$ & $  31.2$ & $  29.3$ & $  28.0$ & $  28.0 $\\
COS-pre & $  34.4$ & $  33.8$ & $  34.4$ & $  34.4$ & $  34.1$ & $  33.8$ & $  34.4$ & $  33.2$ \\
COS-post & $  34.0$ & $  34.8$ & $  35.4$ & $  35.1$ & $  34.4$ & $  34.8$ & $  33.4$ & $  31.4 $\\
\bottomrule
\end{tabular}
    \label{tab:featuresim-10p0}
\end{table}

\begin{table}[h!]
    \small
    \setlength{\tabcolsep}{1pt}   
    \centering
    \caption{Validation performance (mIoU) for teacher fusion over various ranges on Argoverse 2 $2.5\%$ of the labels utilised.} 
\begin{tabular}{lrrrrrrr}
\toprule
Ranges & $  10$ & $  15$ & $  20$ & $  25$ & $  30$ & $  35$ & $  40 $\\
\midrule
Probs. & $  27.7$ & $  27.7$ & $  27.2$ & $  27.6$ & $  28.4$ & $  28.7$ & $  26.8$ \\
Probs.+Thr. & $  25.5$ & $  25.9$ & $  27.3$ & $  28.2$ & $  26.9$ & $  27.5$ & $  26.7$ \\
Feats. & $  24.1$ & $  25.6$ & $  24.8$ & $  23.6$ & $  23.1$ & $  25.8$ & $  26.2$ \\
Feats.+Thr. & $  26.5$ & $  26.6$ & $  27.0$ & $  26.5$ & $  26.1$ & $  26.7$ & $  25.6$ \\
\bottomrule
\end{tabular}
    \label{tab:teacherfusion-2p5}
\end{table}

\begin{table}[h!]
    \small
    \centering
    \caption{Validation performance (mIoU) for teacher fusion over various ranges on Argoverse 2 $10\%$ of the labels utilised.} 
\begin{tabular}{lrrrrrrr}
\toprule
Ranges & $  10$ & $  15$ & $  20$ & $  25$ & $  30$ & $  35$ & $  40$ \\
\midrule
Probs. & $  36.2$ & $  35.5$ & $  36.5$ & $  36.1$ & $  35.5$ & $  35.8$ & $  35.9$ \\
Probs.+Thr. & $  35.9$ & $  36.2$ & $  36.4$ & $  35.4$ & $  35.7$ & $  35.8$ & $  36.0$ \\
Feats. & $  34.0$ & $  34.5$ & $  34.2$ & $  34.4$ & $  34.3$ & $  34.0$ & $  34.4$ \\
Feats.+Thr. & $  34.2$ & $  34.0$ & $  34.8$ & $  34.1$ & $  34.4$ & $  35.4$ & $  34.6 $\\
\bottomrule
\end{tabular}
    \label{tab:teacherfusion-10p0}
\end{table}

\begin{table}[]
    \small
    \centering
    \setlength{\tabcolsep}{4pt}
    \caption{Multi-step Fusion with varying numbers of additional samples within a 20m range using $10\%$ of labels. The best results are achieved with 4 additional samples.}
    \begin{tabular}{l ccc}
    \toprule
     &  \multicolumn{3}{c}{Additional samples} \\ 
     &  $2$    & $4$    & $6$  \\  \midrule
     Probs   & $34.6$ & $35.0$ & $30.2$ \\
     Feats  & $32.4$ & $32.8$ & $26.1$ \\
    \bottomrule
    \end{tabular}
    \label{tab:fusionframes}
\end{table}

\begin{table*}[ht]
    \centering
    \caption{Test performance (mIoU) for multiple label utilisations for both Argoverse 2 (AV2) and nuScenes (nuSc).} 
    \begin{tabular}{lccccccc}
    \toprule
    & +Unlbld &  \multicolumn{6}{c}{Label utilisation} \\ 
    & Data & $2.5\%$ & $5\%$ & $10\%$ & $25\%$ & $50\%$ & $100\%$ \\ \midrule
    \multirow{2}{*}{\rotatebox[origin=c]{90}{AV2}} & \xmark & $7.5^{\pm 0.9}$ & $9.4^{\pm 0.1}$ &  $10.6^{\pm 0.2}$ & $23.4^{\pm 0.5}$ & $33.8^{\pm 0.7}$ & $40.2^{\pm 0.7}$ \\
   & \cmark & $\mathbf{28.9}^{\pm 1.2}$ & $\mathbf{33.8}^{\pm 0.6}$ & $\mathbf{36.7}^{\pm 0.9}$ & $\mathbf{37.4}^{\pm 0.8}$ & $\mathbf{39.6}^{\pm 0.3}$ & $-$ \\ \midrule 
    \multirow{2}{*}{\rotatebox[origin=c]{90}{nuSc}} & \xmark & $5.6^{\pm 0.3}$ & $7.2^{\pm 2.3}$ & $16.6^{\pm 3.0}$ & $24.5^{\pm 2.2}$ & $27.4^{\pm 2.7}$ & $28.6^{\pm 2.4}$ \\
    &\cmark & $\mathbf{15.5}^{\pm 1.2}$ & $\mathbf{22.0}^{\pm 1.6}$ & $\mathbf{25.1}^{\pm 1.9}$ & $\mathbf{27.4}^{\pm 1.8}$ & $27.8^{\pm 2.8}$ & $-$ \\ 
    \bottomrule
    \end{tabular}
    \label{tab:alpha-runs}
\end{table*}

\begin{table*}[ht]
    \centering
    \caption{Performance (mIoU) for city adaptaion for both Argoverse 2 and nuScenes when increasing amount of unlabelled data.} 
    \begin{tabular}{llccccccc}
    \toprule
     &\multirow{2}{*}{Folds} &  \multicolumn{5}{c}{\# unlabelled target sequences} &  Using target \\
     & & $0$ & $25$ & $50$ &$100$ & $200$ & labels \\ \midrule
     \multirow{3}{*}{AV2} & PIT-MIA $\rightarrow$ Rest &   $33.6$&$34.4$&$35.5$&$37.5$&$37.2$&$38.9$ \\
     & MIA-Rest $\rightarrow$ PIT & $36.3$&$38.7$&$40.0$&$40.2$&$40.7$&$41.2$ \\
     & PIT-Rest $\rightarrow$ Mia & $34.5$&$36.1$&$37.5$&$37.8$&$38.5$&$40.2$ \\  \midrule
     nuSc & Boston $\rightarrow$ Singapore & $11.1$ &  $12.6$ & $15.3$ & $17.0$ & $17.7$ & $27.0$ \\
    \bottomrule
    \end{tabular}
    \label{tab:domainadapt}
\end{table*}

\end{document}